\documentclass[11pt,a4paper]{article}
\usepackage[hyperref]{acl2020}
\usepackage{times}
\usepackage{latexsym}
\usepackage{amssymb}
\usepackage{txfonts}
\usepackage{csquotes}
\usepackage{pifont}

\newcommand{\circleone}{\ding{172}~}
\newcommand{\circletwo}{\ding{173}~}

\usepackage{microtype}

\usepackage{graphicx}

\aclfinalcopy 

\usepackage{tabularx}
\newcolumntype{R}[1]{>{\hsize=#1\hsize\raggedleft\arraybackslash}X}%
\newcolumntype{L}[1]{>{\hsize=#1\hsize\raggedright\arraybackslash}X}%
\newcolumntype{C}[1]{>{\hsize=#1\hsize\centering\arraybackslash}X}%

\definecolor{green}{rgb}{0.0, 0.75, 0.0}

\title{Entity Linking via Dual and Cross-Attention Encoders}

\author{Oshin Agarwal\thanks{\enspace Work done during internship at Google} \\
  University of Pennsylvania \\
  \texttt{oagarwal@seas.upenn.edu} \\\And
  Daniel M. Bikel \\
  Google Research \\
  \texttt{dbikel@google.com} \\}

\date{}

\begin{document}
\maketitle
\begin{abstract}

Entity Linking has two main open areas of research: \circleone generate candidate entities without using alias tables and \circletwo generate more contextual representations for both mentions and entities. Recently, a solution has been proposed for the former as a dual-encoder entity retrieval system \cite{gillick-etal-2019-learning} that learns mention and entity representations in the same space, and performs linking by selecting the nearest entity to the mention in this space. In this work, we use this retrieval system solely for generating candidate entities. We then rerank the entities by using a cross-attention encoder over the target mention and each of the candidate entities. Whereas a dual encoder approach forces all information to be contained in the small, fixed set of vector dimensions used to represent mentions and entities, a cross-attention model allows for the use of detailed information (read: features) from the entirety of each $\langle\textrm{mention},\textrm{context},\textrm{candidate entity}\rangle$ tuple. We experiment with features used in the reranker including different ways of incorporating document-level context. We achieve state-of-the-art results on TACKBP-2010 dataset, with 92.05\% accuracy. Furthermore, we show how the rescoring model generalizes well when trained on the larger CoNLL-2003 dataset and evaluated on TACKBP-2010.

\end{abstract}

\section{Introduction}
\label{sec:intro}

Entity linking is the task of finding the unique referring entity in a knowledge base for a mention span in text. For example, \emph{Union City} in the sentence ``The Bay Area Transit Centre in \emph{Union City} is under construction" refers to the Wikipedia entity \texttt{Union\_City,\_California}. Entity linking is typically performed in two steps: generating candidate entities from the knowledge base and then selecting the most likely entity from these candidates. Often, priors and alias tables (a.k.a. candidate tables) are used to generate the set of candidate entities, and work on entity linking has focused on either generating better alias tables or reranking the set of candidate entities generated from alias tables.

Alias tables, however, suffer from many drawbacks. They are based on prior probabilities of a mention referring to an entity using occurrence counts of mention strings. As a result, they are heavily biased towards the most common entity and do not take into account complex features such as mention context and entity description. Such features can help not only in better disambiguation to existing entities but also when new entities are added to the knowledge base without the need for retraining. If candidate generation is based on dense representations instead of a list of candidate ``entity ids", the system will generalize better, even in a zero-shot setting. Lastly, alias tables are often not readily available for many domains, so an approach that is free of alias tables will be useful for such domains. Recently, \cite{gillick-etal-2019-learning}, also known as DEER, used a dual encoder system to learn entity and mention representations in the same space to perform end to end entity linking using nearest neighbor search without the use of alias tables. This model obtained accuracy competitive to the state-of-the-art entity linking systems on the TACKBP-2010 dataset. In this paper, we use this model as a candidate generator by retrieving the top few nearest neighbors.

\begin{figure}[t]
  \includegraphics[scale=0.4,trim={4cm 7.3cm 1cm 4cm},clip]{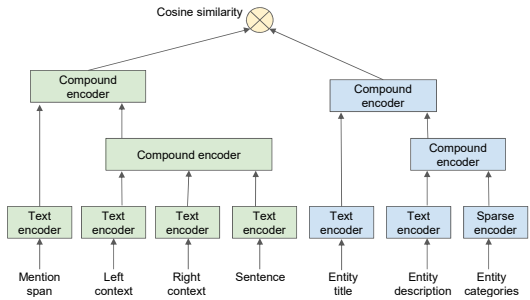}
  \caption{Dual encoder that generates candidate entities for the given mention. Green denotes the mention features and blue denotes the entity features.}
  \label{fig:flare}
\end{figure}

\begin{figure*}
  \includegraphics[width=\textwidth,trim={2cm 5.5cm 1cm 4.7cm},clip]{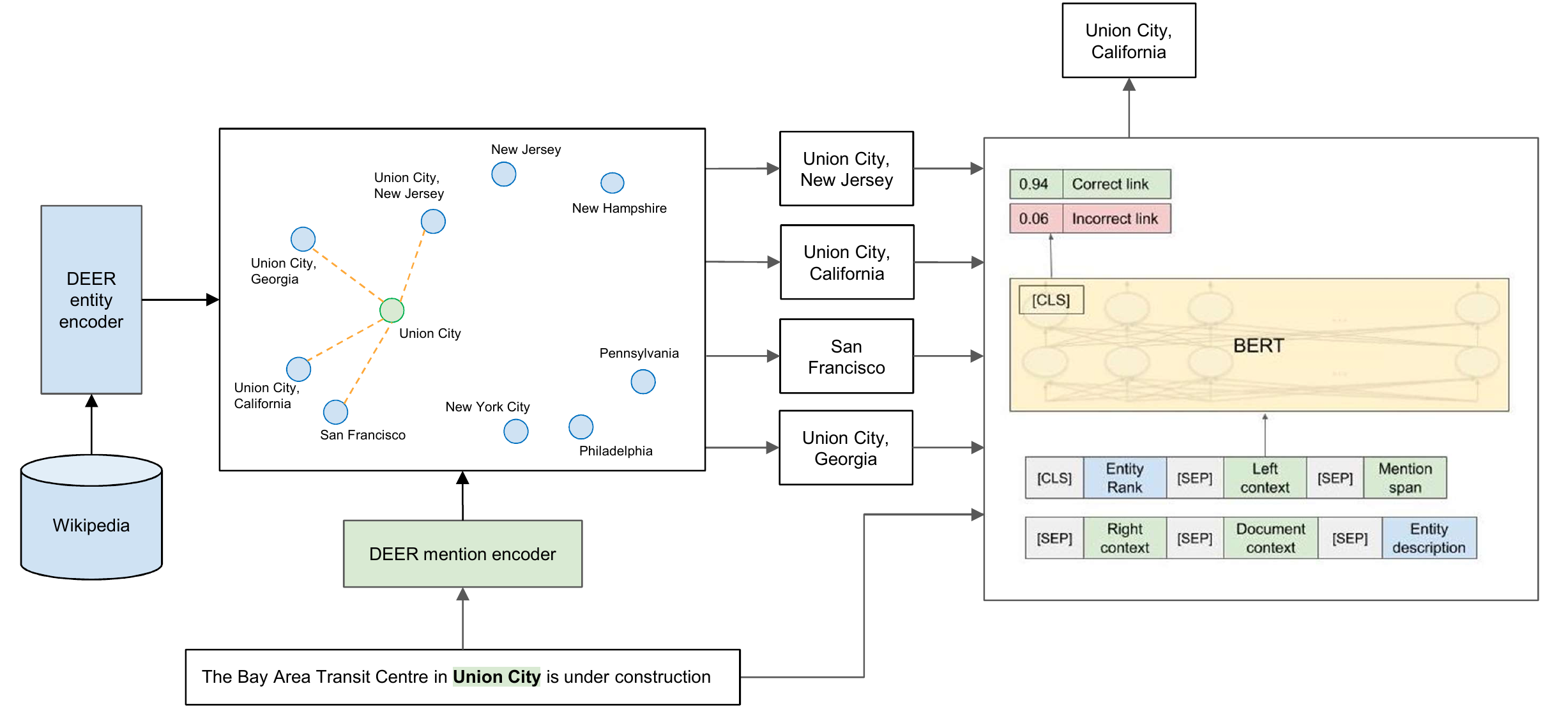}
  \caption{Cross-attention encoder that reranks the candidate entities. Each candidate is paired with the mention and processed one at a time. Green denotes the mention features and blue denotes the entity features.}
  \label{fig:reranker}
\end{figure*}

Another challenge in entity linking is the use of more complex contextual representations for mentions as well as entities. While a dual encoder learns separate fixed representations for entities and mentions in context, we may often want to look at different pieces of information depending on the entity and mention in question. In the above example on \emph{Union City}, if the candidate entity were \texttt{New\_York\_City}, then the names serve as sufficient evidence. If the candidate entity were \texttt{Union\_City,\_New Jersey} though, we need to look at the full sentence with the mention for additional clues. Moreover, depending on the entity, we may even need to look at different parts of the sentence or even the document. On the entity side, we may want to look the relevant parts of the entity description instead of just the title. Let's say we have a mention \emph{Asia Cup} and two candidate entities \texttt{2018\_Asia\_Cup} and \texttt{2016\_Asia\_Cup}. Entity descriptions give us additional information that that the 2018 Asia Cup was held in UAE and the 2016 Asia Cup was held in Bangladesh. Depending on whether the mention context has the year or the location, we would want the entity representation to capture the corresponding relevant information from the entity. For this reason, we re-rank the candidate entities using BERT \cite{devlin-etal-2019-bert} as a cross-attention encoder. Cross-attention gives the opportunity to choose relevant context selectively depending in the specific mention and entity in question and the available features. 

In our approach, the candidate generator uses the lighter representation of GloVe \cite{pennington2014glove} with fewer features as it is trained on the full knowledge base and the reranker uses the more complex representation of BERT and additional features such as document context as it needs to operate only on a small set of candidates and much less data overall. This method gives improved accuracy on both the TACKBP-2010 and CoNLL-2003 datasets. Furthermore, we show that the reranker trained on the larger of the two datasets--CoNLL-2003--generalizes well on both the datasets. We also present an ablation study and a qualitative analysis to determine the impact of various entity and mention features used in the reranker. Lastly, we discuss the challenges faced in the task of entity linking for a fair comparison to prior work and a need to establish standards for a more meaningful comparison of different systems.

\section{Task Setup}
Entity linking involves finding the unique entity in a knowledge base that a mention in text refers to. Given a knowledge base $\mathrm{KB} = \{e_1, e_2 ... , e_n\}$ and a mention in text $m$, the goal is to find the unique entity $e_m \in \mathrm{KB}$ that m refers to. Typically, a two-step procedure is used: \circleone candidate generation and \circletwo reranking candidates. In the first step, possible candidate entities are shortlisted, $\mathrm{CE} = \{e_1, e_2 ... , e_k\}$ s.t. $e_j \in \mathrm{KB}, j: 1 \leq j \leq k$. CE may or may not contain $e_m$. Finally, the entities in CE are re-ranked so that $e_m$ gets selected.

For example, given the mention \emph{Union City} in the sentence ``The Bay Area Transit Centre in \emph{Union City} is under construction", a candidate generator would first generates possible candidate entities from the knowledge base such as \texttt{Union\_City,\_New\_Jersey} and \texttt{Union\_City,\_California}. The reranker would them aim to selects one entity out of these candidates, in this case \texttt{Union\_City,\_California}.

\section{Model}

In this work, we follow a two-step procedure as well: \circleone candidate generation and \circletwo reranking candidates. We use the dual encoder architecture in \cite{gillick-etal-2019-learning} for generating a small set of candidate entities. We use top $k$ retrieved entities as candidates for linking. We rerank these candidates by binary classification using BERT as a cross-attention encoder to generate the probability of each of the candidates being the true link and select the one with the highest probability. We first briefly describe the dual encoder candidate generator, followed by the details of the cross-attention encoder for reranking. The full system architecture is illustrated in Figures \ref{fig:flare}\&\ref{fig:reranker} along with an example. 

\subsection{Candidate Generator}
\label{sec:flare}

We use the dual encoder model in \cite{gillick-etal-2019-learning} for generating candidates for each mention. The model learns representations for entities and mentions in the same space by maximizing the cosine similarity between the two during training. The mention encoder consists of multiple feed-forward networks encoding the mention span, the left context (5 words), the right context (5 words) and the sentence with the mention masked. The entity encoder also consists of multiple feed-forward networks encoding the entity title, description (first paragraph) and user-specified categories. Entity categories are input as sparse vectors. GloVe vectors are used to represent everything else in the input. Wherever the input consists of multiple words, the average of the GloVe representations is taken before feeding into the neural network components. This architecture is shown in Figure \ref{fig:flare}.

The model is trained iteratively on the 2018-10-22 English Wikipedia dump with 5.7M entities and 112.7M linked mentions. The linked mentions serve as positive examples for training and different negatives are selected in every round. The first round uses in-batch random negatives. This is followed by 5 rounds with hard negatives i.e. the top 10 retrieved entities that are more similar to the mention than the true link using cosine similarity.

The retrieval process simply consists of encoding all the 5.7M entities using the entity encoder, then encoding the given mention using the mention encoder and selecting the nearest entity neighbors of the mention using these representations as determined by the cosine similarity. We use the top 100 retrieved entities as candidates for the reranker.

\subsection{Reranker}
\label{sec:reranker}

We model the reranking problem as a binary classification task and fine-tune BERT for the task on domain-specific data. For each pair of mention span and candidate entity, we learn a joint representation of the pair using BERT as a cross-attention encoder. This representation is then classified as true link or not. At test time, the final entity is selected from the candidates as a post-processing step on the ``true link" class probabilities output by the reranking model. That is, for each mention, we select the candidate entity out of the 100 candidates with the highest probability as the final linked entity for that mention.

The input representation to BERT consists of both the mention and entity representation as shown in Figure \ref{fig:reranker}. The mention is demarcated inline in its full sentence using the \texttt{[SEP]} symbol. We also include document-level context for better disambiguation. Document-level context is included as a bag of words in the document that are not present in the mention's local sentence and that are non--stop words. For the entity, we use the entity description and the rank of the candidate entity in the retrieval dual encoder. We do not include the entity name explicitly as the description almost always contains the name in it: e.g., the first sentence for the entity \texttt{India} is ``India \textellipsis is a country in South Asia.'' A different symbol is used to denote each of the candidate entity ranks using an unused token in the BERT vocabulary. For example, \texttt{[unused0]} is used to denote rank 1, \texttt{[unused1]} for rank 2, etc. All the input components are concatenated using the \texttt{[SEP]} symbol and fed as input. 

We experimented with different forms of document context: all other mentions, full document as ordered sequence of tokens, full document as an unordered bag of words. The final model uses document context as an unordered bag of words. While we lose certain multi-word expressions due to this, ordering the document context terms in any way---TF-IDF or the original document order---is inessential to achieving good performance. Having an unordered bag of words, on the other hand, allows it to learn keywords from the document that help identify the entity without the additional ordering constraints. The results from the final model are presented in \S\ref{sec:results} and all other experiments in \S\ref{sec:ablation}.

We use the public BERT large uncased model,\footnote{\texttt{uncased\_L-24\_H-1024\_A-16}} a learning rate of 6e-6, a batch size of 64, 3 epochs and a maximum sequence length of 256. This model performed better than both the cased and the multilingual model.

\section{Results}
\label{sec:results}

We evaluate our model on the test split of TACKBP-2010 dataset. The candidate generator is trained on the 2018-10-22 English Wikipedia dump. The reranker is trained on the training split of TACKBP-2010. The results are presented in Table \ref{table:ComparisonFair}. 

A fair comparison to prior work is a big challenge in entity linking. Wikipedia has been growing rapidly with time, and different systems use different versions of Wikipedia with varying numbers of entities in the knowledge base (KB).\footnote{\url{https://en.wikipedia.org/wiki/Wikipedia:Size_of_Wikipedia}} The larger the set of entities, the more difficult is the task, making comparison to prior work not as meaningful. For this reason, we have two sets of rows in Table \ref{table:ComparisonFair}. The first set of rows have systems that used 800k-1M entities in the KB. For our work, we use the TAC KBP Reference Knowledge Base\footnote{\url{https://catalog.ldc.upenn.edu/LDC2014T16}} for comparison to these systems. TAC KBP has $\sim$818k entities, $\sim$120k of which do not exist in the Wikipedia dump we use. We replace this with random $\sim$120k entities to maintain the same number. We also calculate recall@1 with this smaller set of entities for the dual encoder in \cite{gillick-etal-2019-learning} that we use as a candidate generator. Recall@n accounts for the true link occurring in top n candidates, irrespective of its rank. We obtain an accuracy of 92.05\%, higher than prior state-of-the-art results. For the second set of rows, we select systems that use 5-5.7M entities, which is a more realistic evaluation for the current version of Wikipedia. For our work, we use the same Wikipedia dump used in the candidate generator with $\sim$5.7M entities. We obtain an accuracy of 88.42\%, which is higher or competitive with prior work. While this comparison is somewhat fair, it isn't completely fair as the number of entities isn't exactly the same. For example, \cite{nie2018mention} has a accuracy of 89.10\%, higher than our work but it uses the July 2016 Wikipedia dump that had $\sim$5M entities as opposed to 5.7M entities in one we use.

\begin{table}[t]
\centering
\setlength{\tabcolsep}{2pt}
\begin{tabularx}{\linewidth}{L{0.65}|C{0.35}}
{\bf Model} & {\bf Accuracy (\%)} \\\hline
\cite{sun2015modeling} & 83.90 \\ 
\cite{raiman2018deeptype} & 90.85 \\  
\cite{gillick-etal-2019-learning} & 88.62 \\
This work & 92.05 \\ \hline
\cite{yamada-etal-2017-learning} & 85.20 \\
\cite{nie2018mention} & 89.10 \\ 
\cite{gillick-etal-2019-learning} & 86.86 \\
This work & 88.42 \\
\end{tabularx} 
\caption{Mention-level accuracy on the TACKBP-2010 test set. Wikipedia is growing immensely with time and the difficulty of the task depends on the number of entities in the knowledge base. The first set of rows use 800k-1M entities and the second set of rows use 5-5.7M entities. This allows for a relatively fair comparison, if not perfectly fair as the number of entities is still not exactly the same.}
\label{table:ComparisonFair}
\end{table}

We also evaluate the system using the 5.7M Wikipedia entity set on CoNLL-2003 data. In this case, the reranker was trained on the training split of the CoNLL-2003 data. The results are presented in Table \ref{table:ComparisonFull}. We obtain a huge gain in accuracy by reranking from a 75.71\% recall@1 of the candidate generator to 88.31\% post reranking. However, the recall@100 is 94.04\%  and since we used 100 candidates, there is still scope for improvement via reranking. Similarly for the TACKBP-2010 dataset, the accuracy improves from 86.86\% to 88.42\% on reranking. Again, the recall@100 was 96.27\% leaving room for further improvement.

\begin{table}[t]
\centering
\setlength{\tabcolsep}{2pt}
\begin{tabularx}{\linewidth}{L{0.6}|C{0.25}|C{0.25}|C{0.26}}
{\bf Model} & {\bf CoNLL} & {\bf TAC} & {\bf TAC transfer} \\\hline
DEER Recall@1 & 75.71 & 86.86 & 86.86\\
DEER Recall@100 & 94.04 & 96.27 & 96.27\\
This work & 88.31 & 88.42 & 89.59 \\
\end{tabularx} 
\caption{Mention-level accuracy (\%) on the TACKBP-2010 and CoNLL-2003 test sets. Knowledge base includes 5.7M entities. Reranker is trained on corresponding training splits of the two datasets for the first two columns. The third column is trained on CoNLL and evaluated on TAC.}
\label{table:ComparisonFull}
\end{table}

\begin{table}[t]
\centering
\setlength{\tabcolsep}{2pt}
\begin{tabularx}{\linewidth}{L{0.6}|R{0.35}|R{0.35}|R{0.25}}
{\bf Dataset} & \multicolumn{1}{c|}{\bf Docs} & \multicolumn{1}{c|}{\bf Mentions} & \multicolumn{1}{c}{\bf M/D} \\\hline
CoNLL-03 train & 946 & 18541 & 19.60 \\
CoNLL-03 test & 231 & 4485 & 19.43 \\
TAC-10 train & 1043 & 1059 & 1.01 \\
TAC-10 test & 1013 & 1020 & 1.01 \\
\end{tabularx} 
\caption{Number of documents and mentions in each of the datasets. M/D is the average number of mentions per document.}
\label{table:DatasetStats}
\end{table}

Lastly, we looked the possibility of using a reranker trained on a single, reasonably representative dataset, instead of using domain-specific training data for each of the evaluations. We trained the reranker on CoNLL-2003 training split and evaluated on CoNLL-2003 and TACKBP-2010 test splits. We used CoNLL-2003 for this purpose as it is many times larger than TACKBP-2010 (cf. Table \ref{table:DatasetStats}). The results are presented in first and last column of Table \ref{table:ComparisonFull}. The model trained on CoNLL shows improvements on reranking on the TAC dataset as well. It even achieves higher performance than training on the TACKBP-2010 training split.

In this set-up, we observed one common pattern of errors. In CoNLL, city and country names usually refer to their respective football clubs and the model learned this artifact but in TAC, they refer to the actual city or country. The same generalization was not observed upon training on the TACK training split. The TAC training set is too small and not enough to generalize to other datasets. Future work could involve training the reranker on more general data such as Wikipedia. We chose not to do this for this work because we wanted to keep the data on which the reranker is trained small. Training it on Wikipedia would require some strategy to filter out a reasonably small yet representative set of examples. One strategy could be to filter Wikipedia for the entities that appear in CoNLL training split. We leave such experiments for future work.

The same hyperparameters are used in all the models and were determined using the development (testa) split of CoNLL-2003. We believe that some more improvement could have been obtained on TACKBP-2010 by fine-tuning hyperparameters, but we decided against this as there was no explicit evaluation set to avoid overfitting. 

\begin{table}[t]
\centering
\setlength{\tabcolsep}{2pt}
\begin{tabularx}{\linewidth}{L{0.49}|C{0.1}|C{0.1}|C{0.13}|C{0.12}|C{0.1}|C{0.1}|C{0.3}|C{0.2}}
{\bf Model}   & $m_s$      & $c_l$      & $c_{dm}$     & $c_{db}$     & $e_n$      & $e_d$      & {\bf CoNLL} & {\bf TAC} \\ \hline
DEER R@1      & \multicolumn{6}{c|}{}                                                       & 75.71 & 86.86     \\ \hline
Reranking     & \checkmark &            &            &            & \checkmark &            & 80.69 & 81.76     \\ \hline
Reranking     & \checkmark & \checkmark &            &            & \checkmark &            & 83.34 & 86.27     \\ \hline
Reranking     & \checkmark & \checkmark &            &            & \checkmark & \checkmark & 84.57 & 88.92     \\ \hline
Reranking     & \checkmark & \checkmark & \checkmark &            & \checkmark & \checkmark & 87.12 & -         \\ \hline
Reranking     & \checkmark & \checkmark & \checkmark & \checkmark & \checkmark & \checkmark & 88.31 & 89.59     \\ 
\end{tabularx} 
\caption{Ablation study for input used in the reranker. Reranker is trained on the CoNLL-2003 training split and evaluated on test splits from both CoNLL-2003 and TACKBP-2010. $m_s$ is the mention span, $c_l$ is the local context (sentence), $c_{dm}$ is the document context (other mentions in the document), $c_{db}$ is the document context (document bag of words), $e_n$ is the entity name and $e_d$ is the entity description. $e_n$ usually appears in $e_d$ and isn't included explicitly when $e_d$ is used. Similarly,  $c_{dm}$ is implicitly included in  $c_{db}$.}
\label{table:Ablation}
\end{table}

\begin{table*}[t]
\centering
\small
\setlength{\tabcolsep}{2pt}
\begin{tabularx}{\linewidth}{L{0.12}|L{0.12}|L{0.45}|L{0.35}|L{0.35}}
{\bf Model 1 features} & {\bf Model 2 features} & \multicolumn{1}{c|}{\bf Mention in Context} & \multicolumn{1}{c|}{\bf Model 1 entity} & \multicolumn{1}{c}{\bf Model 2 entity} \\\hline

Top candidate & $m_s$ + $e_n$ & $c_l$ not used - Only for illustration. {\color{red}CZECH} VICE-PM SEES WIDER DEBATE AT PARTY CONGRESS.  & Czech language & Czech Republic \\\hline

$m_s$ + $e_n$  & $m_s$ + $c_l$ + $e_n$ & SOCCER - {\color{red}JAPAN} GET LUCKY WIN, CHINA IN SURPRISE DEFEAT & Japan & Japan national football team \\\hline

$m_s$ + $c_l$ + $e_n$  & $m_s$ + $c_l$ + $e_d$ & CRICKET - {\color{red}LARA} ENDURES ANOTHER MISERABLE DAY & {\em (Lara Dutta)} Lara Dutta is an Indian actress, model and beauty queen who was crowned Miss Intercontinental 1997 and Miss Universe 2000. & {\em (Brian Lara)} Brian Charles Lara is a Trinidadian former international cricketer,[1][2] widely acknowledged as one of the greatest batsmen of all time.) \\\hline

$m_s$ + $c_l$ + $e_d$  & $m_s$ + $c_l$ + $c_d$ + $e_d$ & Police spokeswoman Deborah Denis said Thursday that George Sherryl Whittaker was attacked by scores of honeybees in his yard in the wealthy British dependency's capital, {\color{red}George Town}. & {\em (George Town, Penang)} George Town, the capital city of the Malaysian state of Penang, is located at the north-eastern tip of Penang Island. & {\em (George Town, Cayman Islands)} George Town is a city situated on Grand Cayman island of the Cayman Islands.)  \\
 & & $c_d$: 2008 23:06:18 agitated Cayman swarm year man Wednesday Islands pronounced Firefighters water 74 UTC old bees died attack hoses dead hospital chase attacked 11 shortly 09 & & \\
 
\end{tabularx} 
\caption{Qualitative analysis of reranker with different input features. Reranker is trained on the CoNLL-2003 training split and evaluated on test splits from both CoNLL-2003 and TACKBP-2010. Mention is marked in red in the local context. In each example, model 1 is incorrect and model 2 is correct, illustrating the improvement. $m_s$ is the mention span, $c_l$ is the local context (sentence), $c_d$ is the document context (document bag of words), $e_n$ is the entity name and $e_d$ is the entity description. $e_n$ usually appears in $e_d$ and isn't included explicitly when $e_d$ is used.}
\label{table:Analysis}
\end{table*}

\section{Ablation Study}
\label{sec:ablation}

We perform an ablation study to disentangle the impact of different input mention and entity features to the reranker. We use the model trained on CoNLL-2003 with the full Wikipedia having 5.7M entities. We evaluate this model on both the CoNLL-2003 and TACKBP-2010 test splits. The results for this study are shown in Table \ref{table:Ablation}.

\subsection{Features}
\label{sec:features}

A reranker with minimal features i.e. the mention span and the entity name (cf. row 2) shows an improvement over the top candidate from DEER on CoNLL data. With TACKBP-2010 however, benefits of reranking are observed only when more complex features are employed. Note that the model was trained on CoNLL so disambiguation is harder on TAC due to a domain shift and more complex features encapsulating the context of the mention and the description of the entity are needed.

However, for the reranker, there is a steady improvement in accuracy on both datasets as more complex features are added. 
Incorporating local context for the mention (cf. row 3) leads to an improved accuracy over using just the mention span.
Similarly, using more refined entity features i.e the description for the entity (cf. row 4) instead of just the name leads to improvement as well. The description implicitly includes the name as well in most cases. Furthermore, adding document context for the mention (cf row 5 and 6) helps as well. In row 5, document context includes other mentions marked in the document. Note in Table \ref{table:DatasetStats}, the CoNLL data has many mentions per document whereas the TAC data has on an average only one mention per document so we did not perform this experiment for TAC as there was no additional context to add. Row 6 is our final model from section \S\ref{sec:reranker}. It incorporates document context as a bag of words without stopwords and words in the local sentence.  Possibly keywords from the document help infer the domain or additional subtleties when the domain is same, allowing better resolution. More details on document context follow in the next section.

\subsection{Ordering} 
\label{sec:ordering}

We experimented with ordering the document context, either in the original document order or by TF-IDF based on the training data. Though somewhat counter-intuitive, we saw either no improvement or a small drop in both cases, indicating that the document order is not important and maintaining it possibly forces the model to learn ordering constraints that are not needed, making the task harder. All we need is keywords from the document and unordered bag of words works the best for it.

Since we used document context as an unordered bag of words, we modified BERT to learn only partial positional embeddings for the part of the input that is ordered and none for the document context. However, we observed that this system performed at par with the full system. In fact, the full system was slightly better as BERT uses subwords and the order between these was lost when we removed the positional embeddings from the document context. To investigate this further, we systematically removed positional embeddings one by one from other parts of the input as well that were ordered, \emph{viz.}, the sentence with the mention and the entity description. In both cases, we observed an enormous drop in accuracy from the high 80's to the low 70's. This shows that BERT is successfully able to learn which parts of the input are ordered and unordered, and we can use features such as bag of words input successfully without any modification.

\section{Qualitative Analysis}

Here, we present some examples where adding more and more complex features lead to improvement in the ranker. These are shown in Table \ref{table:Analysis}. In each case, model 2 predicts the entity correctly and model 1 is wrong, illustrating the improvement. Each row compares the better performing model in its previous row with a more complex and an even better performing model, i.e., model 2 in row n becomes model 1 in row (n+1).

\paragraph{simple reranker} Row 1 has a simple reranker based on the mention span and the entity name. The improvement is likely caused due a shift when the reranker is fine-tuned on the domain data. 

\paragraph{span vs sentence} Row 2 compares the models that use the span vs the full sentence for the mention. The one using just the span selects the country \texttt{Japan}, whereas the one with the full sentence selects \texttt{Japan\_nation\_football\_team}, probably because the sentence has words such as `Cricket', `win' and `defeat'.  

\paragraph{entity name vs description} Row 3 gives the example where the models differ in the entity features--name vs. description. The one that looks at just the name incorrectly selects \texttt{Lara\_Dutta}, even though the sentence mentions cricket but one that looks at the description selects \texttt{Brian\_Lara} correctly.

\paragraph{local vs document context} Finally row 5 shows the difference on including the document context. With just the local context, there isn't enough information to say which George Town the mention refers to and \texttt{George\_Town,\_Penang} is selected.
However, the document mentions `Cayman' and `Islands' which gives evidence for selecting the correct entity when the document context is included.

The examples we presented here were mostly the ones with direct strings in the relevant context. Note though that the model uses dense representations and would be able to capture clues in the context far less direct and subtler in nature.

\section{Challenges in Fair Comparison}

Ling et al.~\shortcite{ling-etal-2015-design} describe the many challenges in both the design and evaluation of entity linking systems.  In particular, a fair comparison to prior work is a huge challenge in the task of entity linking due to the ever-changing knowledge base that is Wikipedia. Wikipedia is, in fact, changing rapidly, and previous entity linking efforts have used different versions over time, making comparison difficult at best. We attempted a relatively fair comparison by controlling for the number of number of entities in the knowledge base. However, there are other challenges associated as well that we discuss here.

\paragraph{Number of entities} As shown in section \ref{sec:results}, the number of entities in the knowledge greatly affect the difficult of the resolution task. Different prior works have used different versions of Wikipedia. Some use the TAC-KB which consists of $\sim$818k entities. Others use the full Wikipedia dump that was available whenever the work was conducted. Some others use the 1M most common entities. This leads to some loss in recall on the rare entities. To account for this, some add the entities from the test and development set to the set of 1M, which however makes the task easier. 

\paragraph{Addition of similar entities} Closely related to the number of entities is the addition of new and similar entities. For example, the `Asia Cup' is scheduled every two years and there is an entry in Wikipedia for each of the tournaments so the task of resolving a mention referring to it becomes harder over time. Similarly, the mention `James Love' in the CoNLL data could once be easily disambiguated to the NGO director by this name, whose page was added in 2006. However, Wikipedia now has multiple people by this name. In 2007, the page on James Love, the Kentucky Politician was added and in 2016, the page on James Love, the musician was added, making the task increasingly harder.

\paragraph{Change in entity names} Many entities tend to change names over the course of years. Some of these changes are simple such as `Wal-Mart' being changed to `Walmart', while others are much more drastic such as `Orient-Express Hotels Ltd.' being changed to `Belmond Ltd.'. Unfortunately, entity linking datasets are essentially created once and frozen in time, and are therefore likely to contain old, out-dated entity names as mention spans. For any such dataset containing mentions using old entity names, it is more challenging to disambiguate the old entity mentions when a newer version of Wikipedia is the chosen knowledge base.

There is a need to establish new benchmarks and standards to allow for a fair and controlled comparison of work in the domain. While the TAC KBP Reference Knowledge Base has served as a standard for comparison, it hasn't been universally used in all work and moreover, Wikipedia has grown more than sixfold since, so it is no longer a realistic knowledge base to use in terms of size.

\section{Related Work}
Mihalcea and Csomai~\shortcite{mihalcea2007wikify} developed the seminal Wikify!\ system that used a series of stages to identify important concepts in a document (keywords or phrases) and then to disambiguate those spans by linking them to the appropriate Wikipedia URL, similar to what an author of a new Wikipedia page might do. After identifying key terms/phrases, the system used an alias table to identify possible disambiguations for each keyphrase, making use of a module that measured the term overlap with an entity's definition and a mention's context, as well as a machine learning module based on features of both the mention-in-context and entity.

Similar to the our work, several earlier efforts also use a neural reranker atop a candidate generator \cite{francis-landau-etal-2016-capturing,eshel-etal-2017-named,yamada-etal-2017-learning,gupta-etal-2017-entity,sil-etal-2018-neural}. In particular, Francis-Laundau et al.~\shortcite{francis-landau-etal-2016-capturing} represent one of the first uses of a neural network-based approach for entity candidate scoring, using CNN's to generate fixed-length vectors intended to represent the topics present in each of the mention, its local and document contexts and a candidate entity and its page's text.

Sil et al.~\shortcite{sil-etal-2018-neural} observe that modeling large document-based contexts is expensive. Accordingly, to model a candidate entity's Wikipedia page, they take the tf-idf--weighted average of its word vectors and then train a $\tanh$ activation layer to put that embedding in the same space as the entity mentions to be linked. Furthermore, they employ the output of a coreference resolution system to restrict mention document contexts to be only the sentences in which references of the mention occur.

Gupta et al.~\shortcite{gupta-etal-2017-entity} essentially use a multi-task dual encoder approach, with three different losses for encoding an entity description, its mention contexts and a set of learned representations of entity types obtained from Freebase, all in the same shared embedding space.  In contrast to the DEER model \cite{gillick-etal-2019-learning} we use here for candidate generation, Gupta et al.\ use an explicitly obtained alias table (set of ``mention surfaces'').  This is an important distinction, as in many contexts an alias table is either available but with poor coverage or else unavailable altogether.

Logeswaran et al.~\shortcite{logeswaran-etal-2019-zero} is one of the more directly relevant pieces of work to the present approach. In that paper, the authors attempt zero-shot entity linking on specialized domains---wikias---by using a straightforward BM25-based candidate generator followed by a cross-attention BERT rescoring module trained only on other wikias that could score entities in a target wikia that had only been pretrained on, but not seen during training by the rescorer.  By contrast, we wanted to find out to what extent a cross-attention BERT scoring model could take advantage of a small amount of targeted training, as well as whether a dual encoder could be an effective candidate generator for such an approach.  We also wanted to examine different methods of including document context, including the bag-of-words approach that has proven effective. The model we presented can potentially be used to improve performance in the Wikia dataset as well.

\section{Conclusion and Future Work}
For the entity linking task, we have shown that it is possible to achieve significant accuracy gains over a dual encoder retrieval model by using a BERT rescoring model that can take advantage of cross-attentional features, with state-of-the-art results on the TAC dataset.  Our method involved a fairly straightforward application of BERT as a context-sensitive encoder as the precusor to a classification layer, but fundamentally showing the power it has for this task, particularly when adding additional context.  Furthermore, our experiments revealed that BERT successfully \emph{ignores} positional information when it does not meaningfully contribute to its objective function during training.  Finally, we showed the ability of our model to generalize by training on the CoNLL dataset and testing on TAC.

For future work, we hope to push the dual encoder approach, due to its relatively lower computational cost by simultaneously training the dual encoder and the cross-attention BERT rescorer, adding the rescorer model's loss to the dot product loss for the dual encoder during training.  Our hope is to recover a substantial portion of the performance gains achieved with the cross-attention model within the context of the dual encoder alone, akin to the gains seen with the distillation of BERT models for other tasks.  Within this work, we hope to examine various true ranking models.  At present, we have explored reranking via instance rescoring, which, while effective in practice, does not necessarily leverage additional information that could be brought to bear by a true reranking model that attempts to preserve ``good'' ordering relationships in the lists with which it is presented during training and inference.


\bibliography{entity-linking}
\bibliographystyle{acl_natbib}

\end{document}